\documentclass[runningheads]{llncs}
\usepackage[T1]{fontenc}
\usepackage{graphicx}

\makeatletter
\renewcommand\section{\@startsection {section}{1}{\z@}%
                                   {-12\p@ \@plus -4\p@ \@minus -4\p@}
                                   {12\p@ \@plus 2\p@ \@minus 2\p@}
                                   {\normalfont\large\bfseries}}
\makeatother

\makeatletter
\renewcommand\subsection{\@startsection{subsection}{2}{\z@}%
                                     {-8\p@ \@plus -4\p@ \@minus -4\p@}
                                     {8\p@ \@plus 1\p@ \@minus 1\p@}
                                     {\normalfont\normalsize\bfseries}}
\makeatother
\makeatletter
\renewcommand\subsubsection{\@startsection{subsubsection}{3}{\z@}%
                       {-5\p@ \@plus -4\p@ \@minus -4\p@}%
                       {-0.5em \@plus -0.22em \@minus -0.1em}%
                       {\normalfont\normalsize\bfseries\boldmath}}

\makeatother
\usepackage{hyperref}
\usepackage{color}

\usepackage{algorithm}
\usepackage{algpseudocode}
\begin{document}
\title{Enhancing Explainability in Multimodal Large Language Models Using Ontological Context}
\titlerunning{Enhancing Explainability in MLLM Using Ontological Context}
%
\author{Jihen Amara\inst{1,2}\orcidID{0000-0003-3675-259X}  \and
Birgitta König-Ries\inst{1,2}\orcidID{0000-0002-2382-9722}\and
Sheeba Samuel\inst{3}\orcidID{0000-0002-7981-8504}}
\authorrunning{J. Amara et al.}
%
\institute{Heinz Nixdorf Chair for Distributed Information Systems, Friedrich Schiller University Jena, Jena, Germany \and
Michael Stifel Center Jena, Jena,  Germany  \and
Chemnitz University of Technology, Chemnitz, Germany\\
\email{(jihene.amara, birgitta.koenig-ries)@uni-jena.de\\
sheeba.samuel@informatik.tu-chemnitz.de } \\
}
\maketitle              
\begin{abstract}
Recently, there has been a growing interest in Multimodal Large Language Models (MLLMs) due to their remarkable potential in various tasks integrating different modalities, such as image and text, as well as applications such as image captioning and visual question answering.
However, such models still face challenges in accurately captioning and interpreting specific visual concepts and classes, particularly in domain-specific applications.
We argue that integrating domain knowledge in the form of an ontology can significantly address these issues.
In this work, as a proof of concept, we propose a new framework that combines ontology with MLLMs to classify images of plant diseases.
Our method uses concepts about plant diseases from an existing disease ontology to query MLLMs and extract relevant visual concepts from images. Then, we use the reasoning capabilities of the ontology to classify the disease according to the identified concepts.
Ensuring that the model accurately uses the concepts describing the disease is crucial in domain-specific applications. By employing an ontology, we can assist in verifying this alignment. Additionally, using the ontology's inference capabilities increases transparency, explainability, and trust in the decision-making process while serving as a judge by checking if the annotations of the concepts by MLLMs are aligned with those in the ontology and displaying the rationales behind their errors.
Our framework offers a new direction for synergizing ontologies and MLLMs, supported by an empirical study using different well-known MLLMs.

\end{abstract}
\section{Introduction}

In the last few years, an exciting progress has been made in the field of Natural Language Processing (NLP) with the use of  Large Language Models (LLMs).
These LLMs have shown impressive zero and few shot reasoning  performance on most NLP tasks such as summarization and question answering \cite{xue2020mt5}.
In other words, users can explain a task to the model (prompting) and it will execute it (zero-shot inference), or can offer a few demonstrative examples to guide toward the desired task and output format (few-shot inference)  \cite{doveh2024towards}. Hence, prompting provides a simple method to exploit the capabilities of LLMs without the need for fine tuning.

 Motivated by the success of such models, a huge interest  has been raised toward integrating other modalities such as vision which led to the new field of Multimodal Large Language Model (MLLM).
 One of the best known MLLMs is GPT-4V \cite{openai2023}. Since its release in September 2023, there has been a surge of interest in MLLMs in both research and industry leading to the proposal of various new MLLMs such as LLaVa \cite{liu2024visual} and Gemini \cite{gemini2023}.
Eventhough MLLMs have shown impressive performance, they suffer from various problems such as hallucinations, lack of explainability and domain specific knowledge \cite{pan2306unifying,moradi2021gpt}.
The phenomenon of hallucination arises from the MLLM being trained on non-factual data, thus enabling them to generate highly plausible yet inaccurate responses \cite{minaee2024large,ji2023survey}.
Also, their ability in domain specific knowledge application is still under explored. 
Current MLLMs are primarily trained and tested on general tasks such as visual question answering using common images from the internet which restrict their utility in domain specific fields such as agriculture or healthcare \cite{verma2024mysterious}. 
However, such domains require robust semantic and fine grained understanding abilities. For example in \cite{moradi2021gpt}, the authors showed that these MLLMs are not proficient few-shot learners in the biomedical domain.
In this work, we argue that integrating domain knowledge in the form of an ontology could play an important role in mitigating such problems. 
Hence, in this work, as a proof of concept, we propose a novel framework that combines an ontology with MLLMs to classify images of plant diseases.

Our contributions can be summarized as follows:
\begin{itemize}
    \item \textbf{Semantic enrichment of MLLMs  through ontology driven prompting mechanism:}  
    We propose a method that integrates ontological knowledge into the prompting of MLLM. The ontology provides needed knowledge to guide the model toward the identification of relevant contextual visual concepts.  This will help in improving the model's ability to produce results  that are semantically aligned and coherent with the domain knowledge.
    \item \textbf{Ontologies as  a judge of MLLM performance:}
    Ontologies represent the semantics of a specific domain in a structured way. By evaluating how well the MLLM output aligns with the concepts and relationships defined in the ontology, we can assess how well they understand the domain in study. This not only helps in the evaluation of MLLM but could help in selecting the adequate MLLM for the domain in study. Especially with the fast development of the field, many MLLMs are proposed and choosing the right one is a tedious and time consuming task. Evaluation by ontology could make the selection challenge easier and automatic. Hence, we propose a new evaluation metric for MLLMs based on the ontology.
    \item \textbf{Toward an explainable image classification approach exploiting MLLMs and ontologies:}
     Incorporating the ontology reasoning capabilities into our framework can enhance the explainability of classification. By combining the visual concepts identified by an MLLM with the inferred knowledge from the ontology, the reasoner can generate the classification results through logical inference based on the axioms and relationships defined in the ontology. This approach will make the classification more explainable for the users. It enables them to understand the rationale behind each classification decision. Also, in case of misclassification, this method enables the identification of concepts given by the MLLM that have contributed to the incorrect outcome. 
\end{itemize}
In conclusion, our work contributes to the growing field at the intersection of the semantic web and generative AI.  We offer an approach towards the synergy of ontologies and MLLMs and highlight the opportunities and challenges of such an integration. Also, an empirical study of different MLLMs is presented and the domain of rice plant disease classification is used as a proof of concept. However, our method could be easily extended to other fields where an ontology of visual concepts is present.

\section{Related Work}
There has been a growing interest in the AI-community lately towards  developing different MLLMs \cite{gemini2023,openai2023,liu2024visual}. MLLMs started to be used in multiple tasks such as visual question answering \cite{ishmam2024image},  image captioning \cite{liu2023learning,radford2021learning}, detection \cite{lin2022learning}, segmentation \cite{xia2023gsva} and image generation \cite{ramesh2022hierarchical}.

A standard MLLM architecture is composed of three components which are a pre-trained modality encoder, a pre-trained LLM such GPT-3 \cite{brown2020language} and a modality interface to connect them \cite{yin2023survey}.  An example for a vision encoder is CLIP \cite{radford2021learning} which aligns text with images through pre-training on a massive dataset of paired images and their textual descriptions. 
In \cite{moor2023med}, authors used an MLLM for medical visual question answering (VQA), they pre-trained the model on medical image-text data from publications. Then they evaluated their MLLM's abilities with physicians. They found that their proposed model improves performance in medical VQA by 20 \%.

Even though LLMs and MLLMs have shown impressive performance, they suffer from various problems such as hallucinations, lack of explainability and of domain specific knowledge \cite{pan2306unifying,moradi2021gpt}.
This has inspired the semantic web community to contribute their expertise in mitigating such problems through the incorporation of additional knowledge in the form of ontologies or knowledge graphs (KG).
The survey in \cite{pan2306unifying} provides a summary of such endeavours and outlines future directions toward the synergy of KGs and LLMs.
In \cite{chen2022knowprompt} the authors present a knowledge aware prompt tuning approach where they inject knowledge into the prompt template and employ knowledge as constraints for template optimization. Instead of manual prompts, prompt tuning is based on automatically refining prompts to maximise the performance of LLMs. 
The work in \cite{ye2022ontology} further enhances the knowledge injection method through placing knowledge of entities into the context of LLMs and filtering noisy information. However, their method cannot handle complex or structured knowledge such as OWL reasoning rules and description logic.

Prior to MLLMs, there have been a lot of works that tried to make use of KG for multimodal tasks involving vision and text. This is due to their ability to outline a wide range of explicit class semantics. 

For example \cite{salakhutdinov2011learning} present an approach that makes use of Wordnet \cite{miller1995wordnet} which is a lexical database defining word interrelations, for multiclass object detection problems.
In their approach, hierarchical relationships from WordNet \cite{miller1995wordnet} are used to enable objects with limited training data to leverage statistical strength from related objects.

Other works such as \cite{roy2022improving} exploit commonsense KGs such as ConceptNet \cite{speer2017conceptnet} to improve the connection between visual and semantic embeddings by generating commonsense embeddings and address the task of zero shot classification.
Recent work such as \cite{wang2023fashionklip} introduce FashionKLIP developed for e-commerce which integrates a fashion multimodal knowledge graph to a CLIP model for image text retrieval.
While \cite{ghosal2023language} proposes a new strategy of fusing the language guidance using scene graphs with the pre-trained multi-modal such as CLIP, they found that baseline models such as CLIP can not adequately model complex word knowledge.

Our proposed work for integrating ontology with MLLMs differs from existing approaches in several aspects. First, unlike previous work which focuses on exploiting ontological structure or multimodal capabilities independently, our approach combines both paradigms. By doing so, we aim to exploit their respective strengths to collaboratively tackle image classification tasks,  hence enhancing explainability and deepening our understanding of multimodal performance through rich semantic grounding.
Also, our method takes a step towards an enhanced prompt-based ontology mechanism without the need for huge tuning or optimising efforts.
Moreover, our proposed method makes use of an existing ontology about rice plant diseases without the need of creating a new ontology with multimodal capabilities.
To the best of our knowledge, this is the first approach that tries to combine ontology reasoning and recent MLLMs  and evaluates their strengths and limitations in semantic concept visual identification for a domain specific field such as plant diseases.

\section{Methods}

\begin{figure}
\includegraphics[width=\textwidth]{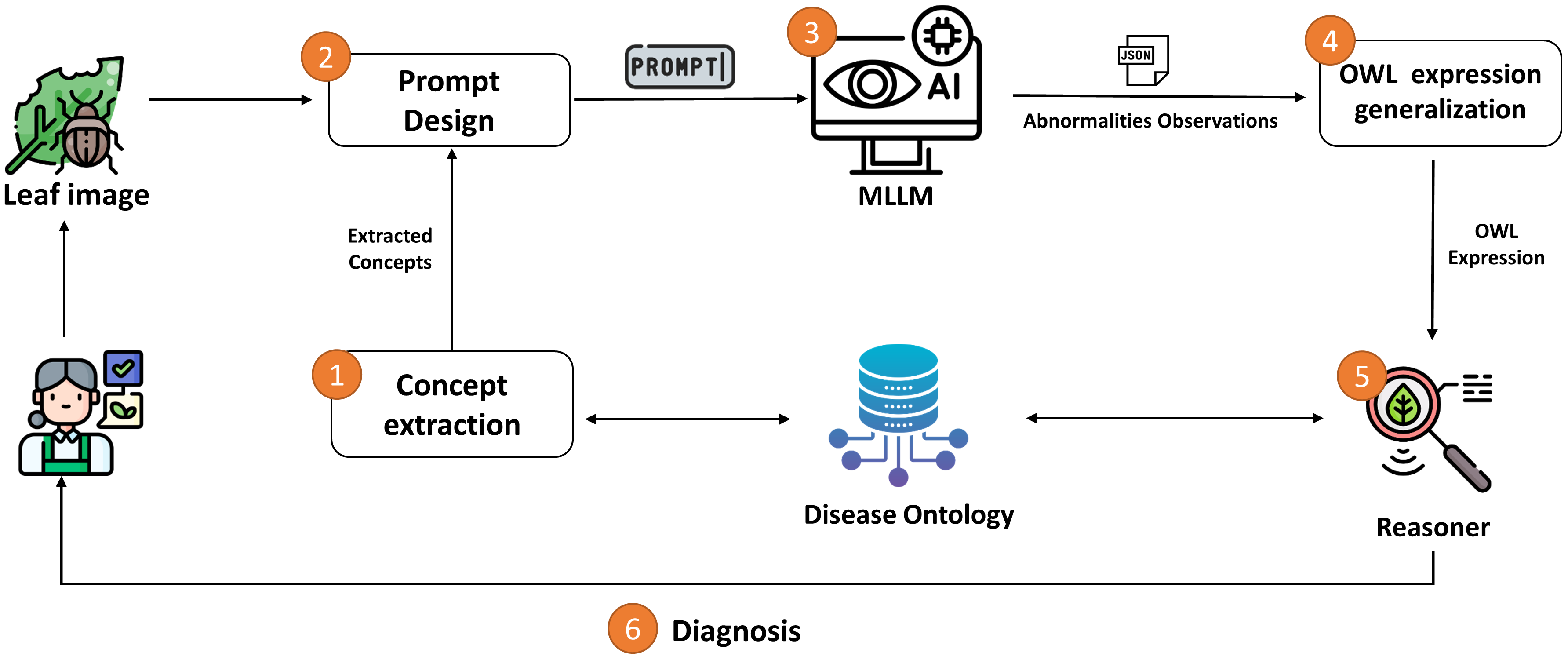}
\caption{The framework of the proposed method.} \label{framework}
\end{figure}
The framework illustrated in Fig.\ref{framework} represents our workflow for integrating ontologies and MLLMs in the context of plant disease classification.

At the core of our approach is an ontology (see Section \ref{sec:domainkg}). 
In the first step (1), concepts representing disease associated abnormalities are extracted from this ontology. These concepts are fed to the prompt designer  (see Section \ref{sec:prompt}). This component leverages them together with an image of a diseased leaf to generate a prompt (2) which is then provided to a MLLM. In its answer, this model returns abnormality observations (3) in a JSON format. The OWL Expression Generator transforms them into a OWL class definition (4). This expression is used by the reasoner to obtain the corresponding disease class from the ontology (5) and return the diagnosis to the user (6) (see Section \ref{sec:query}).

In the upcoming sections, we will describe various parts of our framework in detail. First, we will introduce the used ontology (Section \ref{sec:domainkg}), and then we will explain the prompt design step (Section \ref{sec:prompt}). Finally, we will provide the query construction and reasoning process (Section \ref{sec:query}).

\subsection{Incorporating Domain Knowledge through Ontology Usage} 
\label{sec:domainkg}
In this section, we describe in detail the ontology used for our approach. 
Ontologies play an important role in representing facts and knowledge about a specific domain of interest. 
The abundance of such ontologies motivates us to make use of them in our proposed framework. 
 In this work, we will reuse the RiceDO \footnote{The RiceDO ontology is available in OWL format at \href{https://github.com/RiceManFramework/riceman/blob/master/RiceDO.owl}{https://github.com/RiceManFramework/riceman/blob/master/RiceDO.owl}.} ontology \cite{jearanaiwongkul2021ontology}. It models knowledge related to traits and phenotypes of various rice diseases including abnormal appearance characteristics and symptoms \cite{jearanaiwongkul2021ontology}.
RiceDO was developed taking into account the best practices for ontology design \cite{noy2001ontology} reusing other ontologies such as PDO \cite{jaiswal2017planteome} and PPO \cite{Ammar2009}.
It was developed with the aim of integrating it with an expert system for rice disease identification and disease control recommendations. 
It was evaluated and assessed by ontology experts and senior agronomists, where important criteria such as appropriateness, consistency, and ontology satisfaction were considered \cite{jearanaiwongkul2021ontology}.
Out of the classes contained in the ontology, four are of interest for our case. These are “RiceDisease”, “Abnormality”, “SymptomCharacteristic” and “PlantPart”.

\begin{itemize}
\item \textbf{RiceDisease:}
describes diseases in rice and classifies them into four subclasses: Rice bacterial disease, rice fungal disease, rice phytoplasma disease and rice viral disease. Each disease is then represented as a subclass under the corresponding category (i.e., rice bacterial blight disease is a subclass of rice bacterial disease).
\item \textbf{Abnormality:}
this class was adopted from the PPO \cite{Ammar2009}. When a disease manifests on a leaf it shows a range of observable abnormalities including symptoms such as spots or lesions which indicate the presence of infection, the color of those symptoms (e.g., brown) and the shape of those symptoms (e.g., circular). By carefully observing and analysing such abnormalities, farmers and agronomists can gain the needed information to identify the specific diseases affecting the plant. Hence three subclasses are added to the ontology: ColorAbnormality, SymptomAbnormality and ShapeAbnormality.

\item \textbf{PlantPart:}
Since different symptoms could appear in different plant parts this class is used to define the area of the plant that is infected (e.g., leaf).

\item \textbf{SymptomCharacteristic:}
This class is used to combine frequently used concepts for example, ‘having a spot symptom on leaves’ can be defined by SpotOnLeaf, where SpotOnLeaf represents the definition of a symptom that is ‘(\textit{hasSymptom} some Spot) and (\textit{hasSymptomAt} some Leaf)’.

The object properties we reuse in our approach from RiceDo are the following:  \textit{hasColor}, \textit{hasShapeOfSymptomAbnormality}, \textit{hasSymptom}, \textit{hasSymptomAt} and \textit{abnormalityGroup}. The first three properties are used to define relation from RiceDisease to colorAbnormality, ShapeOfSymptomAbnrmality and SymotomAbnormality, respectively.
 \textit{HasSymptomAt} defines a relation from RiceDisease to PlantPart. Finally, \textit{abnormalityGroup} is used to group various characteristics that can occur together on a certain disease \cite{jearanaiwongkul2021ontology}.
For example, for the rice brown spot fungal disease, the property \textit{abnormalityGroup} is used to group characteristics ‘light yellow halo spot on leaf ’; that is, \textit{abnormalityGroup} some (SpotOnLeaf and (\textit{hasColor} some LightYellow) and (\textbf{hasShape} some Halo)).

\end{itemize}
\subsection{Query Construction and Reasoning }
\label{sec:query}
After getting the abnormality concepts from MLLM, the next step to query the ontology to get the corresponding disease.
RiceDo model rice diseases as TBox axioms using the description logic-based formalism \cite{jearanaiwongkul2021ontology}.

\begin{figure}
\includegraphics[width=\textwidth]{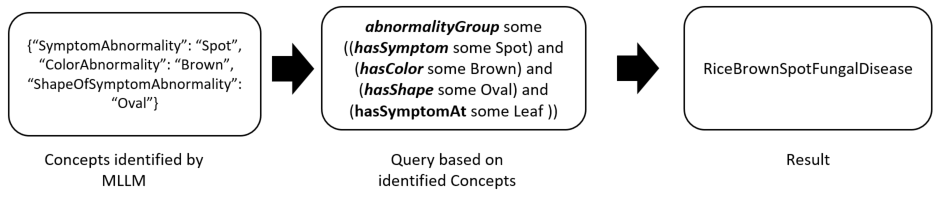}
\caption{Description logic queries for classifying the disease based on the concepts identified by the Multimodal Large Language model (MLLM).} \label{query}
\end{figure}

We employ HermiT reasoner \footnote{\href{http://www.hermit-reasoner.com/}{http://www.hermit-reasoner.com/}.} to identify diseases basesed on subsumption relations between classes.
We implement this using OWL API \cite{horridge2011owl}, a Java API that allows parsing and manipulation of ontological structures and using reasoning engines.
For this aim, a description logic (DL) query is created. The JSON format output from the MLLM is transformed into an OWL class expression using the following steps \cite{jearanaiwongkul2021ontology}:
\begin{itemize}
    \item We align the observed symptom, color and shape using existential quantification via properties \textit{hasSymptom}, \textit{hasColor}, \textit{hasShape} respectively.
    \item We combine these encapsulated expressions through the use of conjunction.
    \item Finally, we associate the full combination using existential quantification through the \textit{abnormalityGroup} property.
\end{itemize}
Figure \ref{query} gives an example of a constructed OWL class expression of the identified visual concepts by MLLM using the previous technique.
\subsection{Prompt design using extracted concepts from the ontology}
\label{sec:prompt}
Prompt engineering is the art of communication with a generative large language model \cite{bsharat2023principled}.
Prompt is defined as the  additional information or hints provided to a model to guide its behavior or help it perform a specific task \cite{gu2023systematic}.
This gives the MLLM the ability to make predictions using only prompts without the need to update their parameters or retrain them on other tasks or datasets. 
Recently different prompt design strategies were proposed \cite{minaee2024large}. However the inclusion of ontology concepts as a  direct contextual information for visual models have not been widely explored as far as we know. Hence in our work, we propose a prompt creation algorithm with ontology context integration.
\begin{algorithm}
\caption{Prompt creation algorithm with ontology context}\label{alg:one}
\begin{algorithmic}
    \State \textbf{Input:} \textbf{Ontology O, \{entity\}} \Comment{In our use case, entity will be "rice leaf"}
    \State \textbf{Output:} \textbf{prompt}
    \State symptom, color, shape $\gets$ ExtractConceptsFromOntology(\textbf{O})
    \State prompt = f"""
As an expert of \{\textbf{entity}\} diseases, your task is to examine the given image of the \{\textbf{entity}\} in a detailed manner to look for color abnormalities, symptom abnormalities, and shape of symptom abnormalities. \\
Alongside the image of \{\textbf{entity}\}, you will be provided with the possible set of color abnormalities and symptom abnormalities and the shape of these symptoms delimited by triple quotes. \\
\Comment{\textbf{Task description}} \\
Return the information in the following JSON format (note xxx is a placeholder, if the information is not available in the image, put “N/A” instead): \\
\{“SymptomAbnormality”: xxx, “ColorAbnormality”: xxx, “ShapeOfSymptomAbnormality”: xxx\} \\
Don't provide anything other than the results in the JSON format. \\
\Comment{\textbf{Output format}} \\
'''
"ColorAbnormality": \{color\}, \\
"SymptomAbnormality": \{symptom\}, \\
"ShapeOfSymptomAbnormality": \{shape\} \\
'''
\Comment{\textbf{Contextual infromation extracted from ontology}} \\
"""
    \State \textbf{return} prompt
\end{algorithmic}
\end{algorithm}

The ontology will be used to help the mapping between the visual level (fine-grained diseases properties) and the semantic level (what is the disease corresponding concepts (i.e., color, symptom, and shape)).

The algorithm \ref{alg:one} takes as input an ontology  $O$  and a given entity that describes the general type of images (e.g., "rice leaf"). First, it starts by extracting relevant concepts from the ontology, focusing on identifying abnormalities. This extraction process involves reasoning and parsing the ontology to locate and isolate abnormalities related to the symptoms, shapes, and colors associated with diseases on the leaf. Once extracted, these concepts are integrated as contextual information for the prompt, enhancing the algorithm's ability to analyze and interpret the image within a well-defined conceptual framework. This ensures a more accurate and relevant analysis, particularly in identifying disease-related abnormalities.

We have also followed the design principles recommended in \cite{bsharat2023principled} specifically the task description, the output format definition and contextual information.

While the prompt created in our study is used in the plant disease classification scenario, the underlying framework is generalizable. For instance, consider the classification of skin diseases. Here, we would have images of skin lesions and an ontology describing various skin diseases based on characteristics such as color, shape, and symptoms. Our prompt structure can be seamlessly adapted to this new context. As shown in algorithm \ref{alg:one}, the prompt can automatically incorporate ontology-derived concepts like colorAbnormality, SymptomAbnormality, and ShapeOfSymptomAbnormality. 
The user would only need to specify which entity is being tested which could also be “skin lesion”.


\section{Experiments and results}
\subsection{Experimental settings}

\begin{figure}
\centering
\includegraphics[width=8cm]{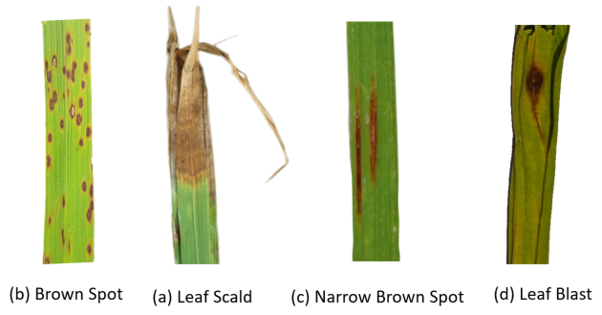}
\caption{Sample images from the  rice leaf disease dataset \cite{hosain2022rice}} \label{dataset}
\end{figure}
Our dataset of rice diseases consists of four distinct classes  which are Brown Spot, Leaf Blast, Leaf Scald and Narrow Brown Spot. For our use case we collected 20 images per disease class. This number is sufficient for our zero-shot classification use case which minimizes the need for a large training dataset. To ensure diversity within our dataset, we employed a dual-source strategy: for each disease class, 10 images were directly collected from \cite{hosain2022rice}, and the remaining 10 images were segmented and extracted from \cite{rice2023}. 

Figure \ref{dataset} presents some sample images from the four disease classes.

\subsubsection{Evaluated MLLMs}

We prompt and evaluate the performance of four leading MLLMs using the ontology: GPT-4V (gpt-4-vision-preview) \cite{openai2023}, Gemini-Pro-Vision (gemini-1.0-pro-vision-001)\cite{gemini2023}, LLaVA (v1.6-7/34b) \cite{liu2024llavanext}, and Claude-3 (opus-20240229) \cite{Claude2023}.
To prompt these models, we used the proposed platform in \cite{yujie2024wildvisionarena}. To have a fair comparison between the MLLMs we use the default parameters in \cite{yujie2024wildvisionarena} where temperature is set to 0.7, Top P is set to 1 and Max output tokens are set to 1024.

\subsubsection{Evaluation metrics}
To measure the degree of alignment between the concepts identified by the MLLM and the ontology defined concepts, we use the Exact Measure ($EM$) metric, where $EM$=1 if the MLLM concept prediction exactly matches the ground truth concept as defined by the ontology otherwise $EM$=0.   
Then, the following accuracies will be measured as follows:
For each visual concept (i.e. symptom, color, and shape) we will count
\begin{equation}\label{eq1}
Concept Wise Accuracy=\frac{True Positive}{True Positive + False Positive}
\end{equation}
where True positive is the number of images where the concept identified  by the MLLM was in alignment with the ontology ($EM$=1); False positive is the number of images where the concept was incorrectly predicted ($EM$=0).
For each image class, we have the concepts from the ontology.

\subsection{Results and Discussion}

\subsubsection{Experiment 1: Ontology as  a judge of MLLM performance}
\paragraph{ \textbf{Multimodal large model evaluation per general concept identification:}}

\begin{figure}
\centering
\includegraphics[width=8cm]{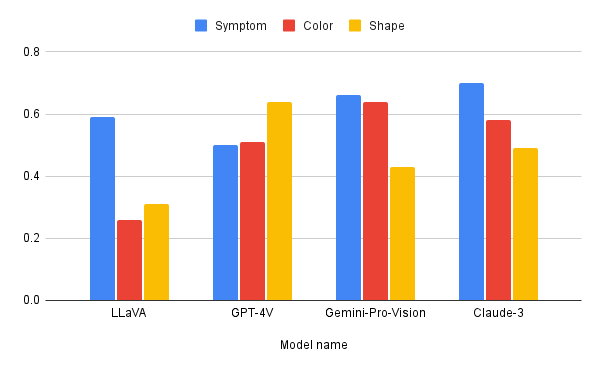}
\caption{Comparative performance of different MLLMs per concept  (i.e, symptom, color and shape) identification.} \label{exp2_results}
\end{figure}

Fig. \ref{exp2_results} shows the performance results  of different MLLMs in fine grained concept identification, particularly focusing on symptoms, colors and shapes. 
For the symptom concept, Claude-3 demonstrates the highest alignment with the ontology concepts, followed by Gemini-Pro-Vision in the second position, while GPT-4V and LLava showed lesser accuracies. This indicates that Claude-3 has a better capacity in capturing symptom-related concepts semantics compared to other MLLMs.
 In addition, the similarity between disease symptoms of different diseases can lead to confusion such as mistaking a spot for a lesion or vice versa. Similarly, for the color concept where Gemini-Pro-Vision showed better performance, GPT-4V and Claude-3 were close in accuracy while LLaVA lags significantly behind. Since we are evaluating the output of MLLMs based on an exact match to the ontology axioms definition of concepts within a specific disease class, we encountered an issue with one example. The LLava model described the color of a spot as “YellowishBrown” whereas the ontology defined it as “brownishYellow”. This difference highlights an interesting aspect for enhancing ontology design in the future by integrating considerations for colors similarities and the models’ perception of colors. However, diseases often depict closely related color symptoms, making it important to capture such small differences for accurate disease classification. Also, in future work, we will adjust accuracy metrics to account for this type of similarity and leverage the similarity between concepts encoded in the ontology. 
Finally, for the shape concept, GPT-4V performed the best compared to the other MLLMs.

\paragraph{\textbf{Multimodal large model evaluation per concept identification for each class:}}

\begin{figure}
\includegraphics[width=\textwidth]{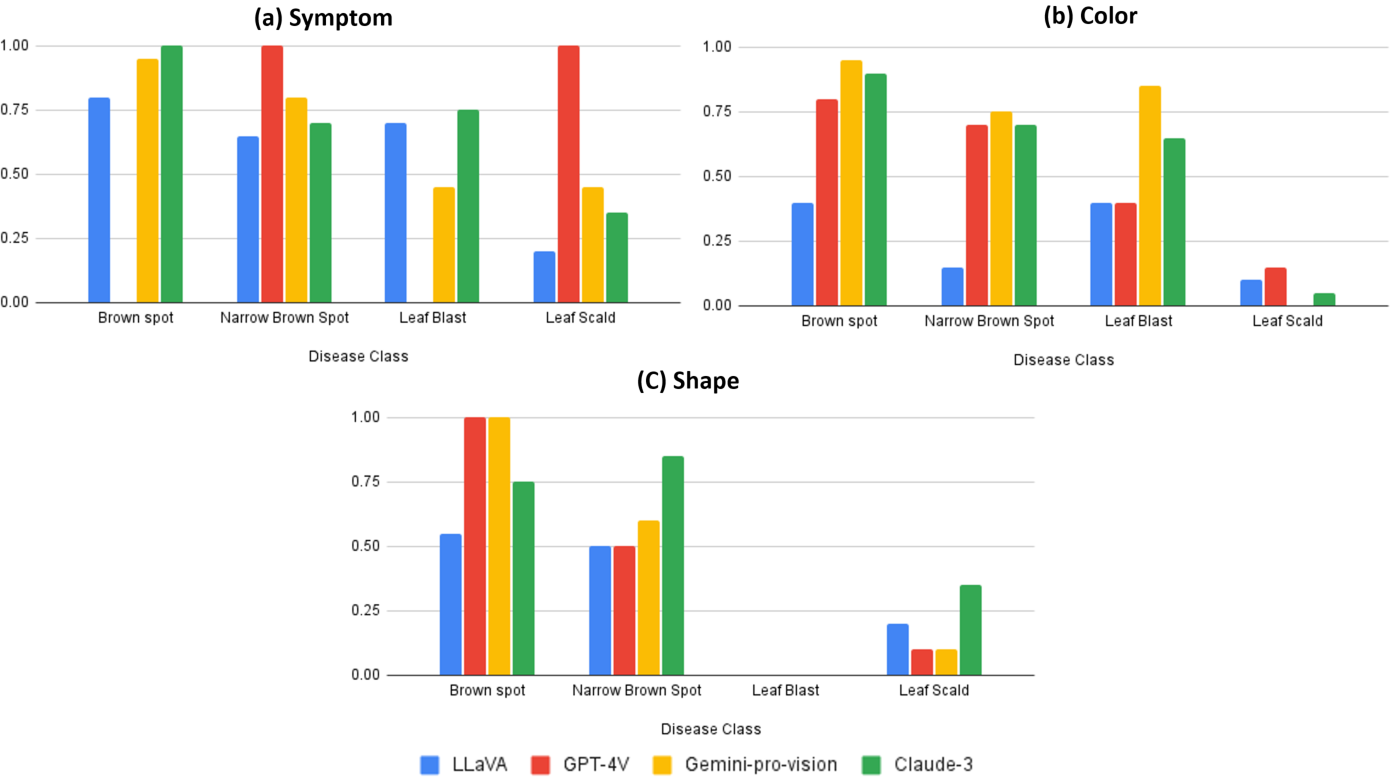}
\caption{Comparative performance of different MLLMs per concept per class.} \label{class_results}
\end{figure}

A second aspect that is important to examine is how well the MLLMs are in identifying specific fine grained semantic concepts within each class. Fig.\ref{class_results} shows this in detail.
In Fig.\ref{class_results}.a, for the Brown Spot, the concept ‘\textit{hasSymptom}.Spot’ was consistently identified by Gemini-Pro-Vision and Claude-3 across all the tested images which shows their alignment with the class definition in the ontology. Also, Llava achieved a comparable performance with an accuracy of 0.8. However, GPT-4V did not accurately detect ‘\textit{hasSymptom}.Spot’  showing a  bias toward interpreting the spot as a lesion.
For the narrow brown spot disease class, the ontology characterises the symptom concept as ‘\textit{hasSymptom}.Lesion’.  While GPT-4V for this class achieved full accuracy, it does not necessarily imply that it accurately identified the concept. Since as we showed above, GPT-4V exhibits a bias toward lesions, potentially affecting its identification. This is also shown in the leaf blast class (‘\textit{hasSymptom}.Spot’) and leaf scald  (‘\textit{hasSymptom}.Lesion’).

In Fig.\ref{class_results}.b, the performance of color concept identification and alignment with ontology per class is presented. As we can see,  GPT-4V, Gemini-Pro-Vision and Claude-3 achieved similar performance levels across Brown Spot, Narrow Brown Spot, while for Leaf Blast class, the performance of GPT-4V was much less than the other two. However, the Scald disease class posed challenges for all three models. On the other hand, LLaVA achieved significantly lower performance across all classes.

 In Fig.\ref{class_results}.c,  the performance of shape concept identification and alignment with concept description by the ontology for each specific class is shown. Brown spot class with its concept ‘\textit{hasShape} some Oval or Circular’ was easily identified by all the models. This is not the case with Narrow Brown Spot where the concept is ‘\textit{hasShape} linear’, where Claude-3 comes first with an accuracy of 0.85 then comes Gemini-Pro-Vision with 0.6 and GPT-4V and LLaVA achieved similar accuracy of 0.5.

None of the models could capture the shape concept for the class Leaf Blast  described by the ontology as ‘ \textit{hasShape} some Eye’, while  most of the models considered it as ‘oval’ shape. 

This highlights a crucial consideration: if we aim to integrate ontologies and MLLMs,  it will be important to thoroughly assess and quantify the similarity of how concepts are identified in both domains. Therefore, refining the embedding space of MLLMs with these concepts could be a significant step toward bringing them closer and enhancing their semantic perception of concepts. 

\begin{figure}
\includegraphics[width=\textwidth]{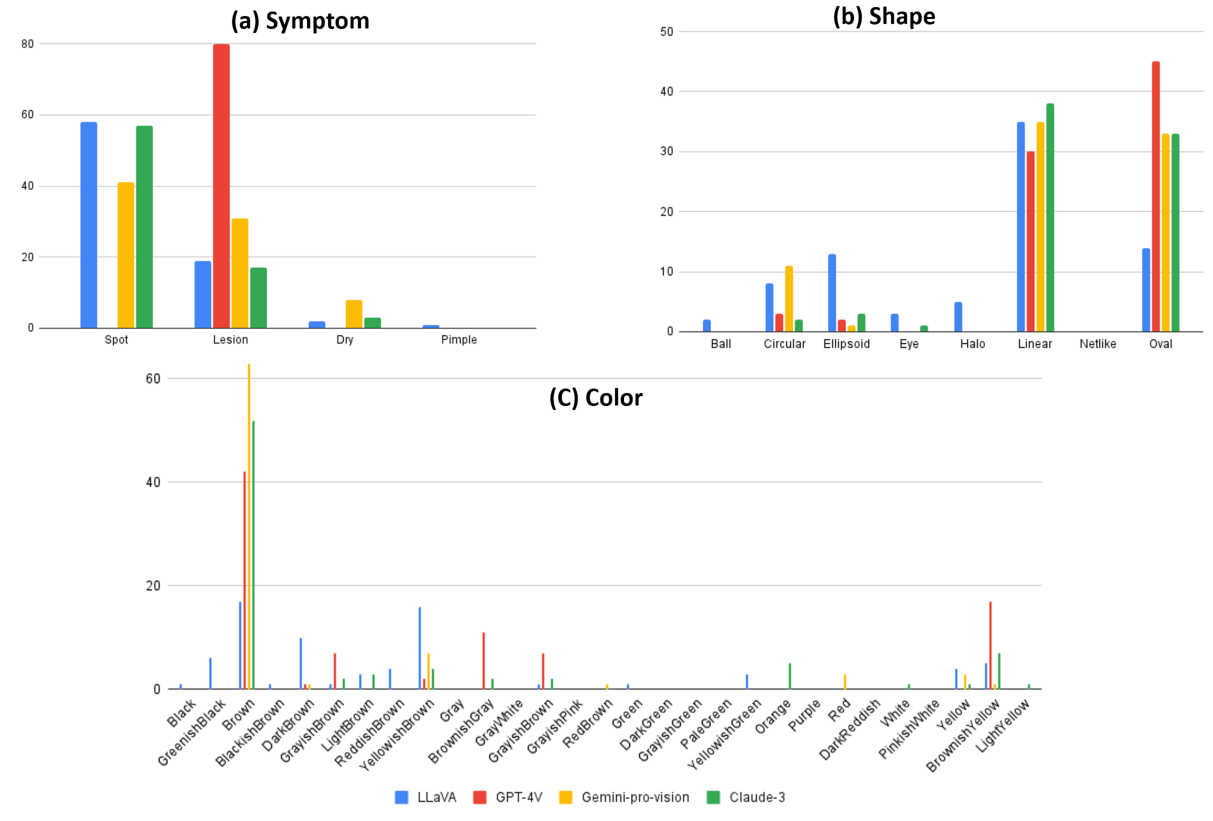}
\caption{Distribution of concepts abnormalities for each  MLLMs.} \label{distribution_results}
\end{figure}

\paragraph{\textbf{Distribution of concepts abnormalities for each model :}}
To have a detailed insight of each model returned answers, we computed the distributions of the different defined ontology concepts for each model.
Figure \ref{distribution_results} shows the occurrence frequencies of different symptoms, shape and colors abnormalities by each model. 
This allows to visually compare the frequency and types of ontology defined abnormalities each model identified, highlighting their sensitivity and specificity in concepts detection tasks.

This alignment helps reveal how closely models' choices are to the expert knowledge which helps in highlighting the areas of strength and the potential bias.

  As we can note in Fig.\ref{distribution_results}.c, brown is  the most frequently chosen color among the set.  Also, in the context of symptoms, the Fig.\ref{distribution_results}.a validate GPT-4V vision bias, that we noticed in previous experiment,  towards classifying symptoms as Lesion. Finally, for shape abnormalities (Fig.\ref{distribution_results}.b), we notice the tendency towards identifying oval shapes. This observation is significant as the ontology often uses oval and circular shapes together  in disease axiom descriptions, suggesting a possible confusion of these categories by the models.
 Such insights are important to ensure that future models are not only accurate but also aligned with expert expectations, leading to more reliable and practical applications.
 
\subsubsection{Experiment 2: Ablation study- prompting healthy leaves with no visual abnormalities concepts}
\begin{figure}
\includegraphics[width=\textwidth]{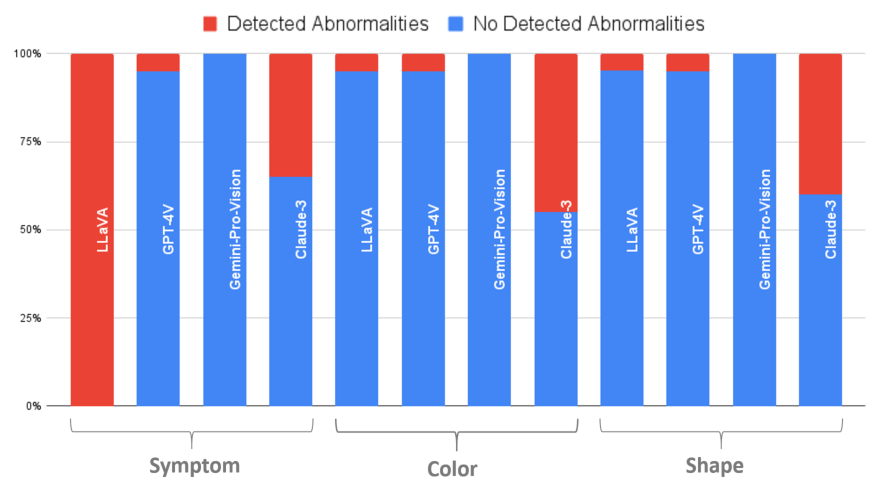}
\caption{Ablation study} \label{ablation}
\end{figure}
We conducted an ablation study with healthy leaves to evaluate the model's ability to adhere to the instructions of "no symptom detected" (i.e N/A) and refrains from reporting abnormalities when none are present. 
Hence, we curated a dataset of 20 images exclusively containing healthy rice leaves. These images were verified to be free from any  abnormalities in symptoms, color or shape.
This helps ensure that the model does not hallucinate symptoms in healthy leaves, confirming that the use of ontological concepts is context-sensitive and appropriate.
Figure \ref{ablation} presents an analytical comparison of the four models: LLaVa, GPT-4V, Gemini-Pro-Vision and Claude 3 evaluated across the three concepts categories abnormalities: Symptom, Color and Shape of Symptom. 
The results indicate a positive outcome, with the majority of the models demonstrating a high percentage of non-detection of abnormalities indicating the efficacy of the approach. 
Specifically, Gemini-Pro-Vision demonstrates 100 \% of no detected abnormalities across all categories which shows its robustness in following the given instructions.
For symptom and shape, if the model returns “N/A” then that’s counted as a no abnormality detected. For color abnormality, if the model return “Green” or “N/A” we consider it as no abnormality detected.  GPT-4V also showed strong performance, with minor exceptions. 
While LLAVA showed strong performance in Color and shape abnormalities, it had a tendency toward hallucinating the symptom despite its non existence.
Finally Calude-3 showed balanced results with higher hallucinations instances than the others. 
These findings show the potential of our proposed approach for the evaluated models while highlighting areas for future improvements for Calude-3 and LLAVA models.

\subsubsection{Experiment 3: Toward an explainable image classification approach exploiting MLLM and ontology}

\begin{figure}
\includegraphics[width=\textwidth]{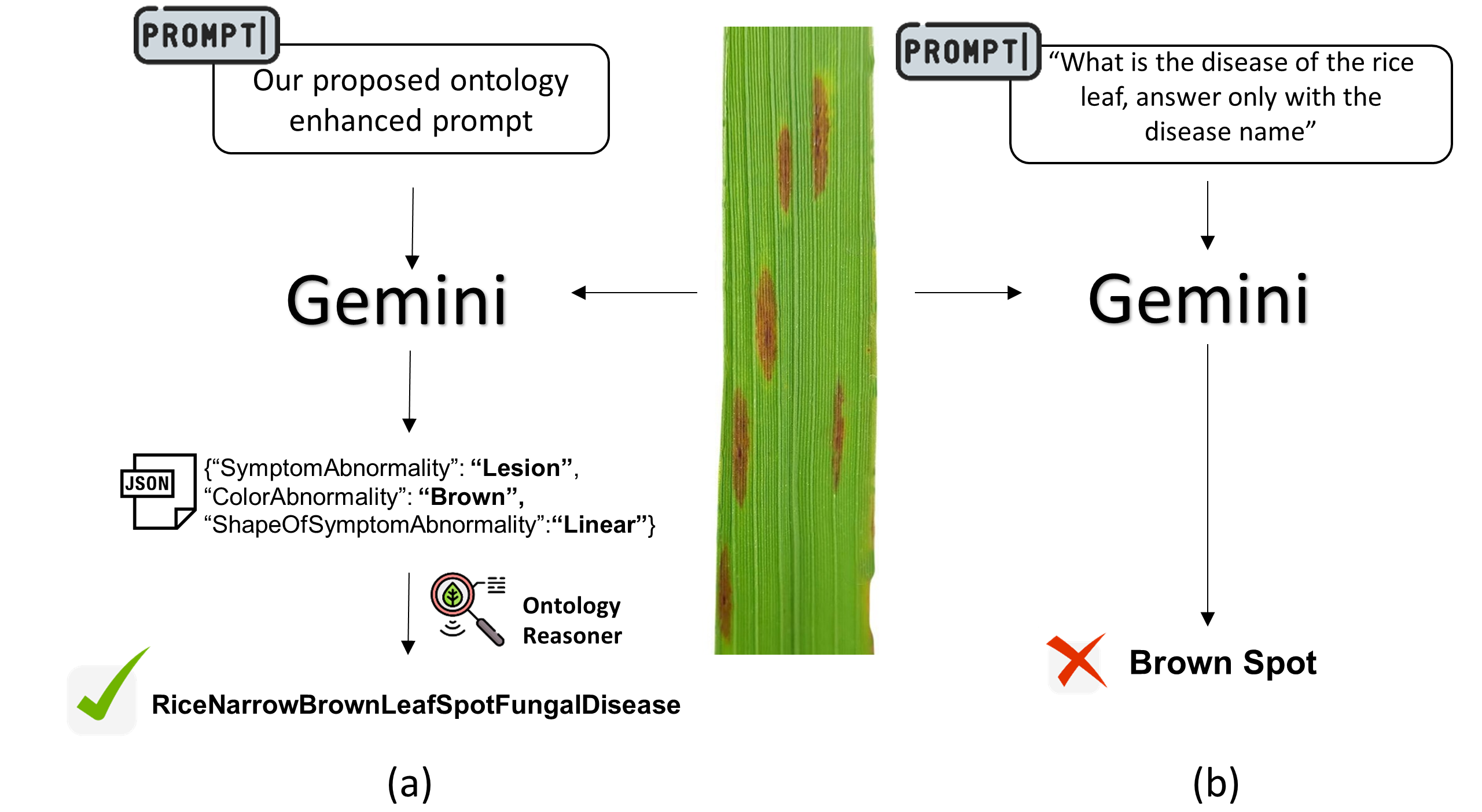}
\caption{Explainability experiment } \label{explain}
\end{figure}

In this experiment, we aim to demonstrate how incorporating ontology reasoning into our framework can significantly improve  the explainability of disease classification compared to  directly prompting an MLLM like Gemini-Pro-Vision.
As shown in Fig.\ref{explain}, we presented Gemini-Pro-Vision with the same image of rice disease twice.
In the first instance, we employed our specifically designed prompt  enriched with relevant ontological knowledge (see Fig.\ref{explain}.a). In the second instance, we used a standard classification prompt as shown in Fig\ref{explain}.b.
When prompted using our proposed approach, Gemini-Pro-Vision accurately identified the disease concepts. Then, by using these identified concepts to reason on the ontology, we were able to achieve a fine-grained classification of Rice Narrow Brown Leaf Spot disease.
In contrast, directly prompting Gemini-Pro-Vision resulted in a misclassification as  brown spot disease.

 This highlights the weakness of MLLMs in distinguishing fine- grained variations between diseases \cite{wu2024comprehensive}, especially when lacking a deeper understanding facilitated by ontological reasoning.

\begin{table}
\caption{Performance of our proposed approach compared to direct prompting using Gemini.}\label{tab1}
\begin{tabular}{|l|l|l|}
\hline
Disease Class &  Direct prompt approach & Our proposed approach\\
\hline
Narrow Brown Spot & 0 & {\bfseries 0.3}\\
Brown Spot & 0.4 & {\bfseries 0.95}\\
Leaf Blast & 0 & 0\\
Leaf Scald & 0 & 0\\
\hline
\end{tabular}
\end{table}

 The performance of our proposed approach compared to direct prompting is detailed in Table \ref{tab1}. Since our method exploits the model’s ability to identify ontology- aligned concepts, its overall performance is still partially dependent on the MLLM inherent capabilities.
However, it successfully classified both Narrow Brown Leaf Spot and Brown Spot diseases.
Furthermore, in cases of misclassification, our approach offers a valuable advantage. It allows us to show the specific concepts identified by the MLLM that contributed to the incorrect outcome. This enables us to explain the reasoning behind the misclassification, such as the model failing to identify the correct semantic symptom (e.g., spot), color (e.g., brown), or shape (e.g., oval) concepts based on the defined ontology. This level of explainability is a significant benefit of our approach.

\section{Conclusion}
The development of Multimodal Large Language Models (MLLMs) represents a significant recent breakthrough in large language models by incorporating multimodality, particularly for understanding visual information.
However, despite their impressive performance, MLLMs suffers from various problems such as hallucinations, lack of explainability and domain-specific knowledge.

In this paper, we contribute to the semantic enrichment and evaluation of MLLMs by incorporating knowledge in the form of an ontology.   
Our framework uses the power of ontologies in many ways: first, by using contextual information extracted from the ontology to prompt different state-of-the-art MLLMs and by using the ontology as a judge to evaluate their capabilities in fine-grained semantic concept identification. Finally, by incorporating ontology reasoning capabilities, our framework provides a new road toward combining both the capabilities of MLLMs and ontology towards an explainable and trustworthy classification.
As a proof of concept for our approach, we worked on the classification problem of plant diseases. However, our proposed framework is general and domain-independent and could be easily extended to other fields.

While our work demonstrates the potential of ontologies with MLLMs, a fundamental limitation remains the difficulty in capturing all expert knowledge within a single ontological framework, which motivates toward combining domain knowledge through the exploration of multiple ontologies.  Also, in future work, we aim to integrate the ontology judgment decision in the learning process to guide the MLLMs toward the most optimal results.
Besides, applying our framework to other fields with a visually defined ontology may help uncover new insights into the MLLM's fine-grained abilities and promote more exploration of the synergy and alignment of these two paradigms.

\begin{credits}

\paragraph*{Supplemental Material Statement:} Our detailed results,
source code and dataset are publicly released as supplemental material on Github
\url{https://github.com/jihenAM/MLLM_ONTO}

\end{credits}
%
%
%
 \bibliographystyle{splncs04}
 \bibliography{mybibliography}

\begin{thebibliography}{10}
\providecommand{\url}[1]{\texttt{#1}}
\providecommand{\urlprefix}{URL }
\providecommand{\doi}[1]{https://doi.org/#1}

\bibitem{gemini2023}
Geminiteam. gemini: A family of highly capable multimodal models, \url{https://blog.google/technology/ai/gemini-api-developers-cloud/}, 2023

\bibitem{openai2023}
Openai. openai models-gpt-4-vision, \url{https://openai.com/research/gpt-4v-system-card}, 2023

\bibitem{Claude2023}
claude.ai, anthropic (2024), \url{https://www.anthropic.com/claude}

\bibitem{Ammar2009}
Ammar, H.: Ontology for plant protection. \url{https://sites.google.com/site/ppontology/home} (2009)

\bibitem{brown2020language}
Brown, T., Mann, B., Ryder, N., Subbiah, M., Kaplan, J.D., Dhariwal, P., Neelakantan, A., Shyam, P., Sastry, G., Askell, A., et~al.: Language models are few-shot learners. Advances in neural information processing systems  \textbf{33},  1877--1901 (2020)

\bibitem{bsharat2023principled}
Bsharat, S.M., Myrzakhan, A., Shen, Z.: Principled instructions are all you need for questioning llama-1/2, gpt-3.5/4. arXiv preprint arXiv:2312.16171  (2023)

\bibitem{chen2022knowprompt}
Chen, X., Zhang, N., Xie, X., Deng, S., Yao, Y., Tan, C., Huang, F., Si, L., Chen, H.: Knowprompt: Knowledge-aware prompt-tuning with synergistic optimization for relation extraction. In: Proceedings of the ACM Web conference 2022. pp. 2778--2788 (2022)

\bibitem{doveh2024towards}
Doveh, S., Perek, S., Mirza, M.J., Alfassy, A., Arbelle, A., Ullman, S., Karlinsky, L.: Towards multimodal in-context learning for vision \& language models. arXiv preprint arXiv:2403.12736  (2024)

\bibitem{ghosal2023language}
Ghosal, D., Majumder, N., Lee, R.K.W., Mihalcea, R., Poria, S.: Language guided visual question answering: Elevate your multimodal language model using knowledge-enriched prompts. arXiv preprint arXiv:2310.20159  (2023)

\bibitem{gu2023systematic}
Gu, J., Han, Z., Chen, S., Beirami, A., He, B., Zhang, G., Liao, R., Qin, Y., Tresp, V., Torr, P.: A systematic survey of prompt engineering on vision-language foundation models. arXiv preprint arXiv:2307.12980  (2023)

\bibitem{rice2023}
Hasan, M., Khatun, S., Raihan, M.A., Uddin, A.H.: Rice leaf bacterial and fungal disease dataset. Mendeley Data, V2 (2023). \doi{10.17632/hx6f852hw4.2}

\bibitem{horridge2011owl}
Horridge, M., Bechhofer, S.: The owl api: A java api for owl ontologies. Semantic web  \textbf{2}(1),  11--21 (2011)

\bibitem{hosain2022rice}
Hosain, A.S., Mehedi, M.H.K., Jerin, T.J., Hossain, M.M., Raja, S.H., Ferdoushi, H., Iqbal, S., Rasel, A.A.: Rice leaf disease detection with transfer learning approach. In: 2022 IEEE International Conference on Artificial Intelligence in Engineering and Technology (IICAIET). pp.~1--6. IEEE (2022)

\bibitem{ishmam2024image}
Ishmam, M.F., Shovon, M.S.H., Mridha, M., Dey, N.: From image to language: A critical analysis of visual question answering (vqa) approaches, challenges, and opportunities. Information Fusion p. 102270 (2024)

\bibitem{jaiswal2017planteome}
Jaiswal, P., Cooper, L., Elser, J., Meier, A., Laporte, M.A., Mungall, C., Smith, B., Johnson, E., Seymour, M., Preece, J.: Planteome: a resource for common reference ontologies and applications for plant biology  (2017)

\bibitem{jearanaiwongkul2021ontology}
Jearanaiwongkul, W., Anutariya, C., Racharak, T., Andres, F.: An ontology-based expert system for rice disease identification and control recommendation. Applied Sciences  \textbf{11}(21),  10450 (2021)

\bibitem{ji2023survey}
Ji, Z., Lee, N., Frieske, R., Yu, T., Su, D., Xu, Y., Ishii, E., Bang, Y.J., Madotto, A., Fung, P.: Survey of hallucination in natural language generation. ACM Computing Surveys  \textbf{55}(12),  1--38 (2023)

\bibitem{lin2022learning}
Lin, C., Sun, P., Jiang, Y., Luo, P., Qu, L., Haffari, G., Yuan, Z., Cai, J.: Learning object-language alignments for open-vocabulary object detection. arXiv preprint arXiv:2211.14843  (2022)

\bibitem{liu2024llavanext}
Liu, H., Li, C., Li, Y., Li, B., Zhang, Y., Shen, S., Lee, Y.J.: Llava-next: Improved reasoning, ocr, and world knowledge (January 2024), \url{https://llava-vl.github.io/blog/2024-01-30-llava-next/}

\bibitem{liu2024visual}
Liu, H., Li, C., Wu, Q., Lee, Y.J.: Visual instruction tuning. Advances in neural information processing systems  \textbf{36} (2024)

\bibitem{liu2023learning}
Liu, H., Son, K., Yang, J., Liu, C., Gao, J., Lee, Y.J., Li, C.: Learning customized visual models with retrieval-augmented knowledge. In: Proceedings of the IEEE/CVF Conference on Computer Vision and Pattern Recognition. pp. 15148--15158 (2023)

\bibitem{yujie2024wildvisionarena}
Lu, Y., Jiang, D., Chen, W., Wang, W., Choi, Y., Lin, B.Y.: Wildvision arena: Benchmarking multimodal llms in the wild (February 2024), \url{https://huggingface.co/spaces/WildVision/vision-arena/}, accessed: March 2024 and June 2024

\bibitem{miller1995wordnet}
Miller, G.A.: Wordnet: a lexical database for english. Communications of the ACM  \textbf{38}(11),  39--41 (1995)

\bibitem{minaee2024large}
Minaee, S., Mikolov, T., Nikzad, N., Chenaghlu, M., Socher, R., Amatriain, X., Gao, J.: Large language models: A survey. arXiv preprint arXiv:2402.06196  (2024)

\bibitem{moor2023med}
Moor, M., Huang, Q., Wu, S., Yasunaga, M., Dalmia, Y., Leskovec, J., Zakka, C., Reis, E.P., Rajpurkar, P.: Med-flamingo: a multimodal medical few-shot learner. In: Machine Learning for Health (ML4H). pp. 353--367. PMLR (2023)

\bibitem{moradi2021gpt}
Moradi, M., Blagec, K., Haberl, F., Samwald, M.: Gpt-3 models are poor few-shot learners in the biomedical domain. arXiv preprint arXiv:2109.02555  (2021)

\bibitem{noy2001ontology}
Noy, N.F., McGuinness, D.L., et~al.: Ontology development 101: A guide to creating your first ontology (2001)

\bibitem{pan2306unifying}
Pan, S., Luo, L., Wang, Y., Chen, C., Wang, J., Wu, X.: Unifying large language models and knowledge graphs: A roadmap, 2023. arXiv preprint arXiv:2306.08302

\bibitem{radford2021learning}
Radford, A., Kim, J.W., Hallacy, C., Ramesh, A., Goh, G., Agarwal, S., Sastry, G., Askell, A., Mishkin, P., Clark, J., et~al.: Learning transferable visual models from natural language supervision. In: International conference on machine learning. pp. 8748--8763. PMLR (2021)

\bibitem{ramesh2022hierarchical}
Ramesh, A., Dhariwal, P., Nichol, A., Chu, C., Chen, M.: Hierarchical text-conditional image generation with clip latents. arXiv preprint arXiv:2204.06125  \textbf{1}(2), ~3 (2022)

\bibitem{roy2022improving}
Roy, A., Ghosal, D., Cambria, E., Majumder, N., Mihalcea, R., Poria, S.: Improving zero-shot learning baselines with commonsense knowledge. Cognitive Computation  \textbf{14}(6),  2212--2222 (2022)

\bibitem{salakhutdinov2011learning}
Salakhutdinov, R., Torralba, A., Tenenbaum, J.: Learning to share visual appearance for multiclass object detection. In: CVPR 2011. pp. 1481--1488. IEEE (2011)

\bibitem{speer2017conceptnet}
Speer, R., Chin, J., Havasi, C.: Conceptnet 5.5: An open multilingual graph of general knowledge. In: Proceedings of the AAAI conference on artificial intelligence. vol.~31 (2017)

\bibitem{verma2024mysterious}
Verma, G., Choi, M., Sharma, K., Watson-Daniels, J., Oh, S., Kumar, S.: Mysterious projections: Multimodal llms gain domain-specific visual capabilities without richer cross-modal projections. arXiv preprint arXiv:2402.16832  (2024)

\bibitem{wang2023fashionklip}
Wang, X., Wang, C., Li, L., Li, Z., Chen, B., Jin, L., Huang, J., Xiao, Y., Gao, M.: Fashionklip: Enhancing e-commerce image-text retrieval with fashion multi-modal conceptual knowledge graph. In: Proceedings of the 61st Annual Meeting of the Association for Computational Linguistics (Volume 5: Industry Track). pp. 149--158 (2023)

\bibitem{wu2024comprehensive}
Wu, T., Ma, K., Liang, J., Yang, Y., Zhang, L.: A comprehensive study of multimodal large language models for image quality assessment. arXiv preprint arXiv:2403.10854  (2024)

\bibitem{xia2023gsva}
Xia, Z., Han, D., Han, Y., Pan, X., Song, S., Huang, G.: Gsva: Generalized segmentation via multimodal large language models. arXiv preprint arXiv:2312.10103  (2023)

\bibitem{xue2020mt5}
Xue, L., Constant, N., Roberts, A., Kale, M., Al-Rfou, R., Siddhant, A., Barua, A., Raffel, C.: mt5: A massively multilingual pre-trained text-to-text transformer. arXiv preprint arXiv:2010.11934  (2020)

\bibitem{ye2022ontology}
Ye, H., Zhang, N., Deng, S., Chen, X., Chen, H., Xiong, F., Chen, X., Chen, H.: Ontology-enhanced prompt-tuning for few-shot learning. In: Proceedings of the ACM Web Conference 2022. pp. 778--787 (2022)

\bibitem{yin2023survey}
Yin, S., Fu, C., Zhao, S., Li, K., Sun, X., Xu, T., Chen, E.: A survey on multimodal large language models. arXiv preprint arXiv:2306.13549  (2023)

\end{thebibliography}

\end{document}